%% file: main.tex
\documentclass[10pt,twocolumn,letterpaper]{article}
\usepackage{cvpr} 
\usepackage{graphicx}
\usepackage{amsmath}
\usepackage{amssymb}
\usepackage{booktabs}
\usepackage[normalem]{ulem}
\usepackage{float} 
\usepackage[pagebackref,breaklinks,colorlinks]{hyperref}

\usepackage[capitalize]{cleveref}
\crefname{section}{Sec.}{Secs.}
\Crefname{section}{Section}{Sections}
\Crefname{table}{Table}{Tables}
\crefname{table}{Tab.}{Tabs.}

\newcommand\blfootnote[1]{%
  \begingroup
  \renewcommand\thefootnote{}\footnote{#1}%
  \addtocounter{footnote}{-1}%
  \endgroup
}

\usepackage{lipsum}
\usepackage{xcolor}
\usepackage{makecell}
\usepackage{bbding}

\AtBeginDocument{%
  }
    
\newcommand{\myname}{TANGLED }

\begin{document}

\def\name{\myname: Generating 3D Hair Strands from Images with Arbitrary Styles and Viewpoints}

\title{\name}

\author{Pengyu Long$^{1,2}$\thanks{Equal contribution.}
\quad
Zijun Zhao$^{1,2*}$
\quad
Min Ouyang$^{1,2*}$
\quad
Qingcheng Zhao$^{1,2}$
\quad\\
Qixuan Zhang$^{1,2}$\thanks{Project Leader.}
\quad
Wei Yang$^{3}$
\quad
Lan Xu$^{1}$
\quad
Jingyi Yu$^{1}$
\\
$^{1}$ShanghaiTech University \quad $^{2}$Deemos Technology \\ $^{3}$Huazhong University of Science and Technology 
\\
{\tt\small \{longpy, zhaozj2022, ouyangmin2022, zhaoqch1, zhangqx1, xulan1, yujingyi\}@shanghaitech.edu.cn} \quad\\ {\tt\small weiyangcs@hust.edu.cn}
\and \small\url{https://sites.google.com/view/tangled1}
}
\twocolumn[{%
\renewcommand\twocolumn[1][]{#1}%
\maketitle
\vspace{-28pt}
\begin{center}
    \centering
    \captionsetup{type=figure}
    \includegraphics[width=\textwidth]{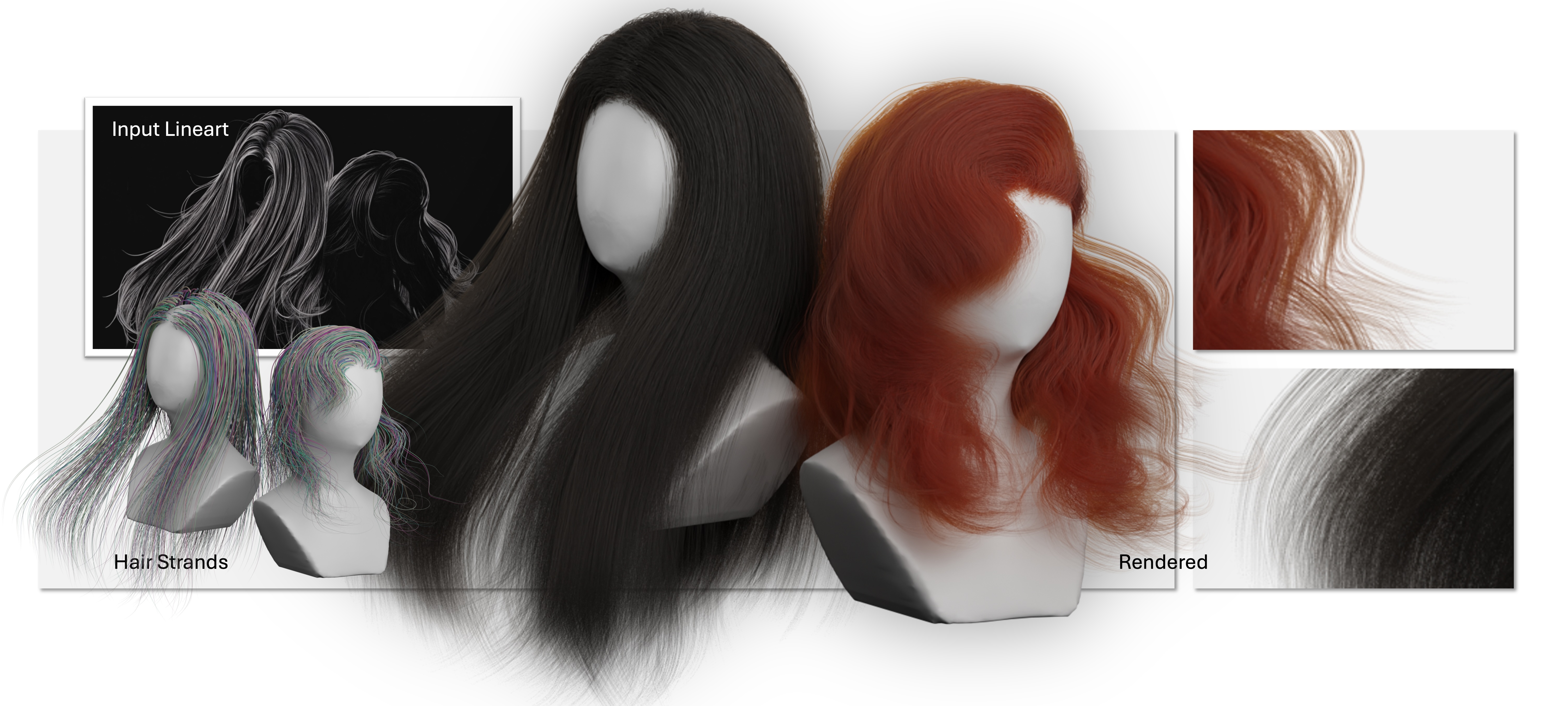}
    \captionof{figure}{TANGLED brings creativity to life by generating high-quality 3D hairstyles from images of any style or viewpoint, seamlessly integrating into existing CG pipelines and delivering breathtakingly detailed hair assets.}
\end{center}%

}]

\blfootnote{$^*$ Equal contribution.\ \ $\dagger$ Project Leader.}
\begin{abstract}

Hairstyles are intricate and culturally significant with various geometries, textures, and structures. Existing text or image-guided generation methods fail to handle the richness and complexity of diverse styles. We present TANGLED, a novel approach for 3D hair strand generation that accommodates diverse image inputs across styles, viewpoints, and quantities of input views.
TANGLED employs a three-step pipeline. First, our MultiHair Dataset provides 457 diverse hairstyles annotated with 74 attributes, emphasizing complex and culturally significant styles to improve model generalization. Second, we propose a diffusion framework conditioned on multi-view linearts that can capture topological cues (e.g., strand density and parting lines) while filtering out noise. By leveraging a latent diffusion model with cross-attention on lineart features, our method achieves flexible and robust 3D hair generation across diverse input conditions. Third, a parametric post-processing module enforces braid-specific constraints to maintain coherence in complex structures.
This framework not only advances hairstyle realism and diversity but also enables culturally inclusive digital avatars and novel applications like sketch-based 3D strand editing for animation and augmented reality. 

\end{abstract}
\input{TANGLED/sections/1_introduction}

\input{TANGLED/sections/2_related_works}

\input{TANGLED/sections/3_dataset}

\input{TANGLED/sections/4_method}
\input{TANGLED/sections/5_results}

\input{TANGLED/sections/6_conclusion}

\input{TANGLED/sections/7_figures}
\bibliographystyle{ieee_fullnames}
\bibliography{main}

\end{document}

%% file: TANGLED/sections/1_introduction.tex
\section{Introduction}
\label{sec:intro}

Hairstyles are powerful symbols of cultural identity, social status, and personal expression. Beyond their aesthetic appeal, they embody deep historical and social meanings. For instance, traditional Asian hairstyles often carry spiritual and societal significance, while African hairstyles like dreadlocks have been central to expressions of cultural pride and resistance. Successfully recreating these hairstyles in digital form, whether in feature films, computer animations, or video games, is vital for fostering inclusive and diverse representation in media. However, realistic hairstyles exhibit intricate geometry, varied textures, and dynamic movement, often requiring artists to painstakingly model each strand. Even with advanced tools, this process remains labor-intensive, demanding a balance of technical expertise and cultural sensitivity to ensure the hair appears authentic and meaningful.

Instead of relying solely on manual modeling, researchers have explored 3D reconstruction solutions to recover hairstyles directly from images. Multi-view capture systems, such as multi-camera domes or moving-camera setups, coupled with reconstruction techniques from early stereo matching methods to modern volumetric modeling approaches, can approximate the overall shape of hair. Yet, these methods struggle to capture the diversity of intricate styles. For example, dreadlocks, with their intertwining locks and varying thicknesses, present significant challenges due to their dense, coiled geometry, which is difficult for 3D scanning systems to recover with fine detail and precision. Moreover, multi-view capture requires specialized equipment or controlled setups, limiting its practicality for casual users or scenarios without access to such apparatus. While 3D capture technologies can handle basic shapes, they fall short in reproducing the richness, detail, and cultural nuances of more complex or specific hairstyles.

Generative tools like DALL·E~\cite{betker2023dalle3}, DreamStudio~\cite{dreamstudio2025}, and Artbreeder~\cite{artbreeder2025} have been developed to create diverse 2D images including hairstyles using text prompts and the latest efforts have been focused on extending this capability to 3D ~\cite{zhang2024clay, rombach2022high, Deitke2023objaverse, Wu2023OmniObject}. Generating 3D hair from text requires bridging the gap between abstract textual descriptions and the detailed geometry, textures, and physical behavior of hair. Text-Conditioned Generative Model~\cite{HAAR:CVPR:2024}, represents a significant step forward in leveraging textual descriptions for synthesizing 3D hair strands. However, these approaches are fundamentally constrained by the accuracy of textual annotations available in datasets. 

In place of texts, it’d be more natural to use images as inputs. Image-based hair modeling methods~\cite{kuang2022deepmvshair,wu2024monohair,zhou2018hairnet,wu2022neuralhdhair,zheng2023hairstep,he2024perm} rely on optimization-based reconstruction techniques, taking single-view or multi-view images as input to refine occupancy and orientation fields. While these techniques can produce high-quality results, they rely on photorealistic images and are computationally expensive, demanding significant time and resources. Moreover, due to the scarcity of diverse hairstyles, particularly afro or curly hair, in commonly used datasets, their effectiveness in capturing such complex structures remains limited. Finally, compounded by sparse training data, they exhibit limited geometric fidelity when modeling intricate structures like braids, which demand precise topological constraints.  

In this paper, we introduce \textit{TANGLED}, a novel approach to generating 3D hair strands from flexible image inputs with diverse styles, viewpoints, and varying numbers of views. We adopt a trilogy to enable such unique flexibility.
First, we introduce the MultiHair Dataset, a curated collection of 457 hairstyles spanning 74 global and local hair attributes (e.g., strand styling, length, direction, layering) with multi-view image annotations. Prior datasets~\cite{hu2015uschairsalon, shen2023ct2hair, Hair20k, zhou2018hairnet} disproportionately represent 10-400 hairstyles, whereas MultiHair prioritizes underrepresented textures (e.g., coiled, locs) and complex geometries, expanding hairstyle diversity by 30\%. This shift tackles the scarcity of diverse training data that undermines generalization in existing methods.

Second, we propose a diffusion framework conditioned on multi-view linearts for flexible hair generation. As a sparse structural representation, such lineart not only preserves topological cues (e.g., parting lines, strand density) but also filters out noise such as lighting variations and occlusions. As a result, lineart conditioning can effectively handle geometric ambiguities inherent in single-view inputs while generalizing seamlessly across diverse image styles and viewpoints. 
Specifically, we adopt a latent diffusion model with the polyline-based representation of 3D hair strands~\cite{HAAR:CVPR:2024} in the latent space. We apply DINOv2~\cite{oquab2024dinov2} on the lineart images to obtain lineart features, which are adapted into the diffusion model via cross attention. We further enhance the generalization ability by randomly blending lineart features from different viewpoints. As a result, regardless of their style or viewpoint, our framework empowers flexible and accurate 3D hair strand generation that adapts to a wide variety of input conditions
Finally, we design a parametric post-processing module that inpaints braid-specific constraints (e.g., cyclic strand crossings, torsion) during generation. It significantly reduces geometric distortions, preserving coherent braid appearance compared to pure diffusion or parametric baselines~\cite{zeng2024hairdiffusion}.

By supporting underrepresented styles like braids and locs~\cite{meishvili2024hairmony}, TANGLED enables culturally inclusive digital avatars. It extends to applications such as sketch-based 3D strand editing, enabling expedited virtual prototyping for animation and augmented reality, where rapid, user-guided design is paramount.Experiments demonstrate superior performance: user studies report 84.3\% preference for our results over text-guided models~\cite{HAAR:CVPR:2024} in realism and diversity.

%% file: TANGLED/sections/2_related_works.tex
\section{Related Work}
\label{sec:related}

Our work bridges advancements in 3D hair representation, image-based hair modeling, and data-driven hair generation.

\vspace{4pt}
\noindent \textit{3D Hair Representation.}
Early methods relied on parametric representations for structured hair geometry design, including 2D parametric surfaces~\cite{koh2000real,liang2003enhanced,noble2004modelling}, wisp-based models~\cite{watanabe1992trigonal}, generalized cylinders~\cite{chen1999system,choe2005statistical,patrick2004modelling,xu2001v,yang2000cluster}, multi-resolution cylinders~\cite{kim2002interactive,wang2004hair}, and hair meshes~\cite{yuksel2009hair,bhokare2024real}. While these enabled intuitive styling, they struggled to capture intricate geometric variations (e.g., curls, frizz) or dynamic properties (e.g., wind interactions) in complex hairstyles. Recent volumetric approaches, such as adaptive shells~\cite{wang2023adaptive}, prioritize efficiency over fidelity.
Strand-based representations emerged as the new standard for high-fidelity modeling. Widely adopted in research~\cite{piuze2011generalized,shen2023ct2hair} and production~\cite{chiang2015practical,fascione2018path}, they excel in physics-based simulation~\cite{daviet2023interactive,fei2017multi,hsu2023sag,herrera2024augmented,digitalsalon} and strand-level editing~\cite{xing2019hairbrush}. However, manually creating a full strand-based hairstyle is highly labor-intensive, highlighting the need for generative models to automate hair asset creation.
Recent advances~\cite{HAAR:CVPR:2024} map strands to UV-space latents. While generative models trained in this space enable hair generation from text prompts, they struggle with limited 3D hair dataset diversity.
Our work addresses the issue by proposing a diverse 3D hair dataset featuring stylized and cultural hairstyles, and a lineart-conditioned generative framework. This enables rapid prototyping of diverse, stylized hair geometries.

\vspace{4pt}
\noindent \textit{Image Based Hair Modeling.}
Early image-based hair modeling methods focused on direct reconstruction~\cite{kong1998generation} or heuristic volumetric techniques~\cite{paris2004capture} from multi-view images. Subsequent work prioritized reconstructing 3D orientation fields to triangulate strands. ~\cite{nam2019strand} introduced a line-based PatchMatch MVS algorithm for robust oriented point clouds, where~\cite{takimoto2024dr} enhanced orientation consistency via global optimization.~\cite{zhou2024groomcap} further advanced fidelity using neural implicit representations. However, occlusion limits multi-view methods to surface-level geometry, producing incomplete internal strands.
Data-driven methods mitigate this by integrating hair structure priors. ~\cite{kuang2022deepmvshair,wu2024monohair} infer internal 3D orientations via neural networks, and~\cite{zakharov2025human,sklyarova2023neural} employ diffusion priors with SDS loss~\cite{poole2022dreamfusion} to refine geometry. Despite progress, these methods require controlled capture setups.

Single-view modeling is a more challenging task. Early methods~\cite{chai2016autohair,chai2015high,hu2015single,chai2012single,chai2013dynamic} relied on database retrieval and refinement, while ~\cite{zhou2018hairnet} predicted 3D models from 2D orientation maps. Volumetric techniques~\cite{saito20183d,zhang2019hair} reconstructed orientation/occupancy fields.
~\cite{wu2022neuralhdhair} improves resolution via a coarse-to-fine framework. HairStep~\cite{zheng2023hairstep} addressed this with depth-strand map hybrid representations, better linking 2D inputs to 3D geometry. 
%
DeepSketchHair~\cite{shen2020deepsketchhair} used sketches to generate 3D orientation fields, bypassing photorealism.
Our work supports multi-style inputs (sketches, stylized art, photos) and resolves ambiguities in sparse data with lineart-conditioned diffusion, enabling robust hair generation across artistic domains.

\vspace{4pt}
\noindent \textit{Hair Generation Models.}
Despite progress in hair modeling, data-driven hair generation remains underdeveloped with notable challenges. GroomGen~\cite{zhou2023groomgen} introduced the first generative model for 3D hair, using a hierarchical representation with strand-VAE and hairstyle-VAE to encode individual strands and overall hairstyles. 
Perm~\cite{he2024perm} proposed a PCA-based parametric model that disentangles hair structures with frequency-domain strand representations. It supports single-image inputs by projecting 3D hairstyles into 2D and aligning them via perm parameter adjustments. 
Recently, HAAR~\cite{HAAR:CVPR:2024} introduced the first text-guided generative framework for 3D hair using a latent diffusion model conditioned on text inputs in a unified hairstyle UV space. 
The hair generation field lacks diverse, high-quality datasets representing a wide range of hairstyles. Additionally, text-based approaches struggle to capture intricate geometric and structural variations in hairstyles.
Our method leverages lineart conditioning and the MultiHair dataset for hair generation. Supporting various input styles (e.g., sketches, photos, partial renders), our approach enhances flexibility, precision, and controllability, bridging user intent and algorithmic output.

%% file: TANGLED/sections/3_dataset.tex
\section{Dataset Construction}
\label{sec:data}
Training a neural model to generate diverse 3D hairstyles requires a dataset that represents ethnic and cultural diversity, ensuring realistic strands between populations.

Moreover, conditioning annotations should be designed to guide the generation process, providing the model with clear, structured information about hair characteristics.

Existing public datasets lack sufficient diversity in hairstyles and detailed annotations. To address this, we first collected a multi-modal 3D hairstyle dataset, $MultiHair$, which includes a variety of hairstyles paired with text and image annotations, as shown in Tab.~\ref{tab:comparison_data}.

\vspace{8pt}
\noindent \textit{Data Collection.}
Advances in hair modeling have been driven by the USC-HairSalon dataset~\cite{hu2015uschairsalon}, which covers conventional styles like bob, afro, curly, and wavy. However, it lacks a detailed representation of complex braided styles, such as ponytails, pigtails, and various braids, due to challenges in data collection and annotation. While studies~\cite{HAAR:CVPR:2024, he2024perm, Hair20k} used augmentation techniques (e.g., squeezing, stretching, cutting, flipping, curling) to enhance diversity, intricate braided styles remain underrepresented.
Thus, we introduce a dataset that broadens hairstyle diversity, emphasizing braided styles. It includes hairstyle strands sourced from various online platforms, reflecting global and local characteristics~\cite{meishvili2024hairmony}. Notably, 10\% of the dataset focuses on braided styles, such as ponytails, pigtails, and diverse braid patterns. To ensure quality, we collaborated with professional artists who refined the data through cutting, stretching, flipping, and blending while preserving the original hairstyles. As shown in Fig.~\ref{tab:comparison_data}, our dataset surpasses existing public datasets in hairstyle diversity and braided style representation.

\begin{table}[tbp]

  \caption{\textbf{Statistical Comparison of Hair Datasets }. We present a comparative analysis of our MultiHair Dataset against existing public datasets across five key dimensions: hairstyles (H) / categories of hairstyles (C), total number of individual hairstyles, availability of text annotations, the number of viewpoints, and image annotations. For open-source datasets, we report cleaned/verified data quantities, while closed-source datasets reflect claimed values. Please note HiSa\&HiDa collects portrait images instead of 3d hairstyles, HairNet provides synthetic 4-view renders of each hairstyle.)}
  \label{tab:comparison_data}

    \resizebox{0.95\columnwidth}{!}{
    \begin{tabular}{lcccc}
    \toprule
             & Basis & Total & Text Annotation & Image Views \\ 
    \midrule
    USC-HairSalon               & 343 (H)   & 343       &  \XSolidBrush & ---              \\
    Hair20k                     & 343 (H)   & 3715      &  \XSolidBrush & ---          \\
    CT2Hair                     & 10 (H)    & 10        &  \XSolidBrush & ---          \\
    GroomGen                    & 35 (C)    & 7712      &  \XSolidBrush & ---         \\
    HAAR                        & 393 (H)   & 9825      &  \Checkmark   & ---          \\
    HiSa\&HiDa                  & 1250*     & 1250*     &  \XSolidBrush & 1         \\
    HairNet                     & 340 (H)   & 40k+      &  \XSolidBrush & 4*         \\
    \textbf{MultiHair (Ours)} & \textbf{457 (H)}   & 10274     &  \Checkmark   & \textbf{72}           \\
    \bottomrule
    \end{tabular}
    }
    \vspace{-7pt}

\end{table}

\begin{figure}[tbp] 
  \includegraphics[width=\columnwidth]{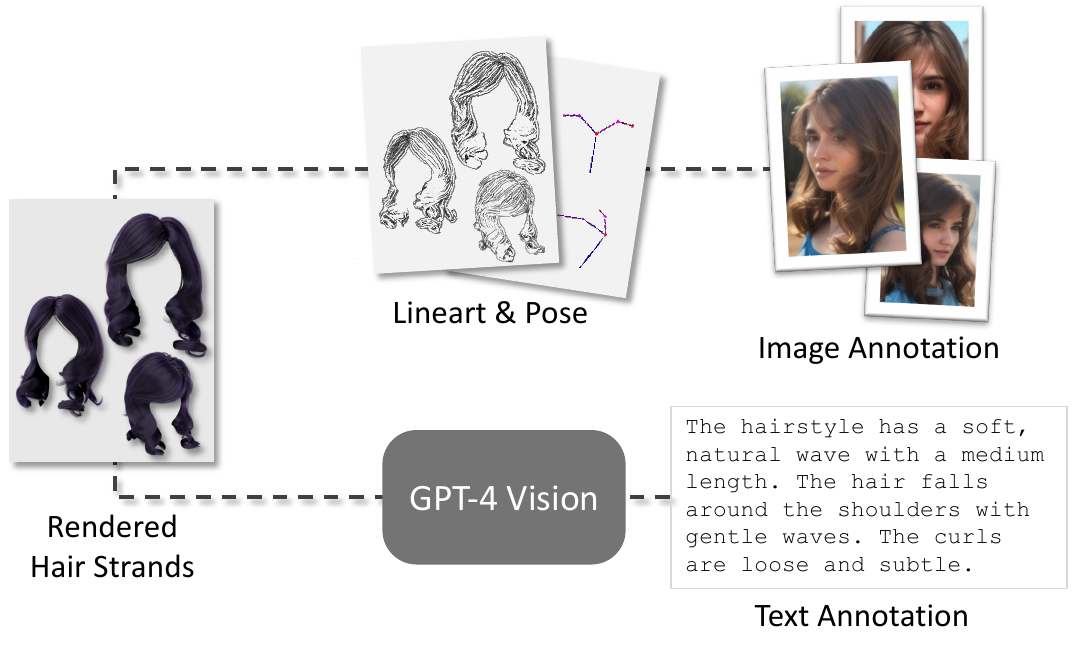}
  \caption{\textbf{Dataset Annotation Process}. 
  Our annotation pipeline begins by processing rendered 3D hair strands with a line-art detector, and line-art sketches are combined with OpenPose~\cite{cao2017realtime} skeletal data for conditioning ControlNet. To enrich dataset diversity, we further synthesize multi-view images, to cover variations in lighting, texture, and perspective. Finally, GPT-4~\cite{chatgpt2025} generates detailed textual annotations for each hairstyle, including attributes such as length, curliness, density, and cultural style.
  }
  \label{fig:DataAc}
\end{figure}

\vspace{8pt}
\noindent \textit{Image and Text Annotation.}
Hairstyles exhibit significant diversity, making them hard to describe accurately solely with text. This complexity stems from factors like texture, shape, and cultural significance, which differ across styles. While images provide intuitive representations, they are often impacted by noise, artifacts, and extraneous factors that obscure details. A robust representation is needed to capture texture, shape, and style while minimizing background clutter, lighting issues, and occlusions. We propose using lineart as an effective representation, focusing on outlines and contours to emphasize shape and structure while avoiding intricate textural or tonal details.
To encompass a broad range of styles and viewpoints, we enhance our dataset by generating diverse hair images with various viewpoints for each 3D hair model.
Specifically, we use eight evenly spaced horizontal viewpoints and three pitch angles. For each combination of viewpoint and pitch angle, we render images with three focal lengths: 35 mm wide-angle, 50 mm standard, and 85 mm telephoto. We then extract the lineart image of each rendered view using a Lineart Detector~\cite{zhang2023adding}. To cover a wide range of hair images and styles, we generated 72 stylized images using ControlNet~\cite{zhang2023adding} conditioned on viewpoints and linearts, to augment the image and lineart annotations, as described in Fig.~\ref{fig:DataAc}.

\begin{figure*}[tbp]
  \centering
  \includegraphics[width=\textwidth]{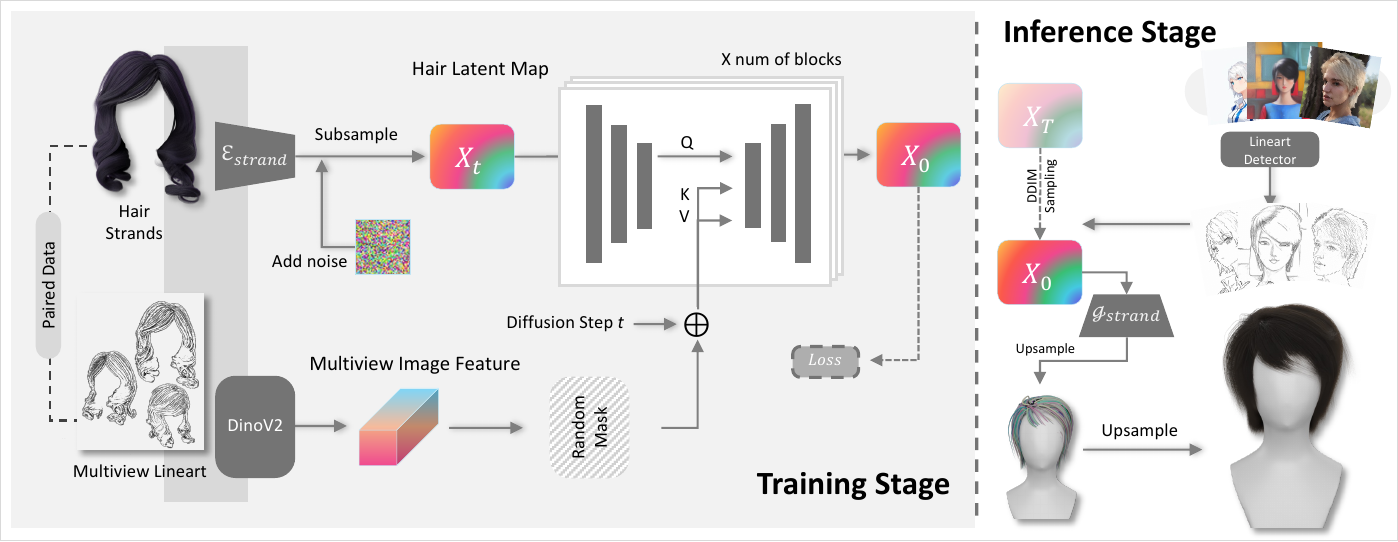}
  \captionof{figure}{\textbf{Architecture of our \myname.} Our model takes hair images with arbitrary styles and viewpoints as conditions, and generate the 3D hair latent through the diffusion process. The conditions are randomly masked and cross-attention with the latent. At inference, we sample hair latent maps and feed the upsampled hair latent map to the strand decoder to extract the 3D hair strands.
} \label{fig_overview}
\end{figure*}

%% file: TANGLED/sections/4_method.tex
\section{The TANGLED Model}
\label{sec:method}

As shown in Fig.~\ref{fig_overview}, given a hair image from an arbitrary viewpoint, \myname 
generates 3D hair strands represented as polylines, making them compatible with standard computer graphics pipelines.

Its lightweight architecture ensures efficient generation of CG-ready 3D hairstyles, which integrate seamlessly into tools like Blender and Unreal Engine for animation and rendering.

\subsection{Hair Strands Latent Diffusion Model}

We adopt the representation of 3D hair strands as a set of 3D polylines evenly distributed over the scalp, as introduced by HAAR~\cite{HAAR:CVPR:2024} and Neural Haircut~\cite{sklyarova2023neural}. These 3D polylines are encoded into a 2D latent map $\mathbf{X}$ utilizing a VAE encoder with scalp UV mapping. 
Specifically, each hair strand $S_i = \{s_j\}, {j \in 1, \dots, L}$ is a sequence of 3D points $s_j$ of length $L$, Each strand is encoded into a latent vector $z$ using an encoder $\mathcal{E}$ and then reconstructed through a decoder $\mathcal{D}$. 
\begin{align}
\mathbf{X} =  \mathcal{M}_\text{uv} (\mathcal{E}(\{ S_j \})), \, \, \, 
S_j = \mathcal{D} \big (  \mathcal{N}_\text{uv} ( \mathbf{X}, j) \big ),
\end{align}    
where $\mathcal{M}_\text{uv}$ denotes the mapping of a strand latent to the scalp UV space, and $\mathcal{N}_\text{uv}$ indicates the process of sampling latent in the UV space.
During training, the VAE is optimized through the L-2 distance between predicted points $\hat{s}_j$ and ground-truth points $s_j$, the cosine similarity between predicted directions $\hat{d}_j = \hat{s}_{j + 1} - \hat{s}_{j}$ and ground-truth directions $d_j$, and the L-2 distance between predicted curvatures $\hat{g}_j = \|\hat{d}_j \times \hat{d}_{j+1}\|_2$ and ground-truth curvatures $g_j$. The fidelity term is expressed as:
\begin{equation}    
\mathcal{L}_{\text{dist}} = \sum_{j, \, S} \|\hat{s}_j - s_j\| + (1 - \hat{d}_j \cdot d_j) + \|\hat{g}_j - g_j\|,
\end{equation}

weight of each term is omitted for clarity. 

The training objective then is the combination of the fidelity term and a KL divergence term.

The denoising process is learned on the latent map $\mathbf{X}$. To optimize computational efficiency, we downsample the latent map to a $32 \times 32$ resolution. The diffusion model adopts a 2D U-Net architecture for denoising within this downsampled latent space. During training, Gaussian noise is progressively added to $\mathbf{X}$ at each time step, resulting in a noisy latent map $\mathbf{X}_t = \mathbf{X} + \mathbf{n}_t$, where $\mathbf{n}_t$ represents the Gaussian noise at time step $t$. The model $\epsilon{\theta}$ is then trained to predict and remove this noise, effectively learn the reverse diffusion process that denoises the latent map over time.

\begin{equation}    
\epsilon_\theta(\mathbf{X}_t; t, \mathbf{F}) = \mathbf{n}_t
\end{equation}

where $\mathbf{F}$ represents the concatenated muti-view lineart features extracted by DINOv2~\cite{oquab2024dinov2} for conditioning. The model is trained by minimizing the Mean Squared Error (MSE) loss between the predicted noise and the noise added at each diffusion step.

\subsection{Multi-view Lineart Conditioning}
\label{sec:image_conditioning}

\begin{figure}[tbp]
  \includegraphics[width=\columnwidth]{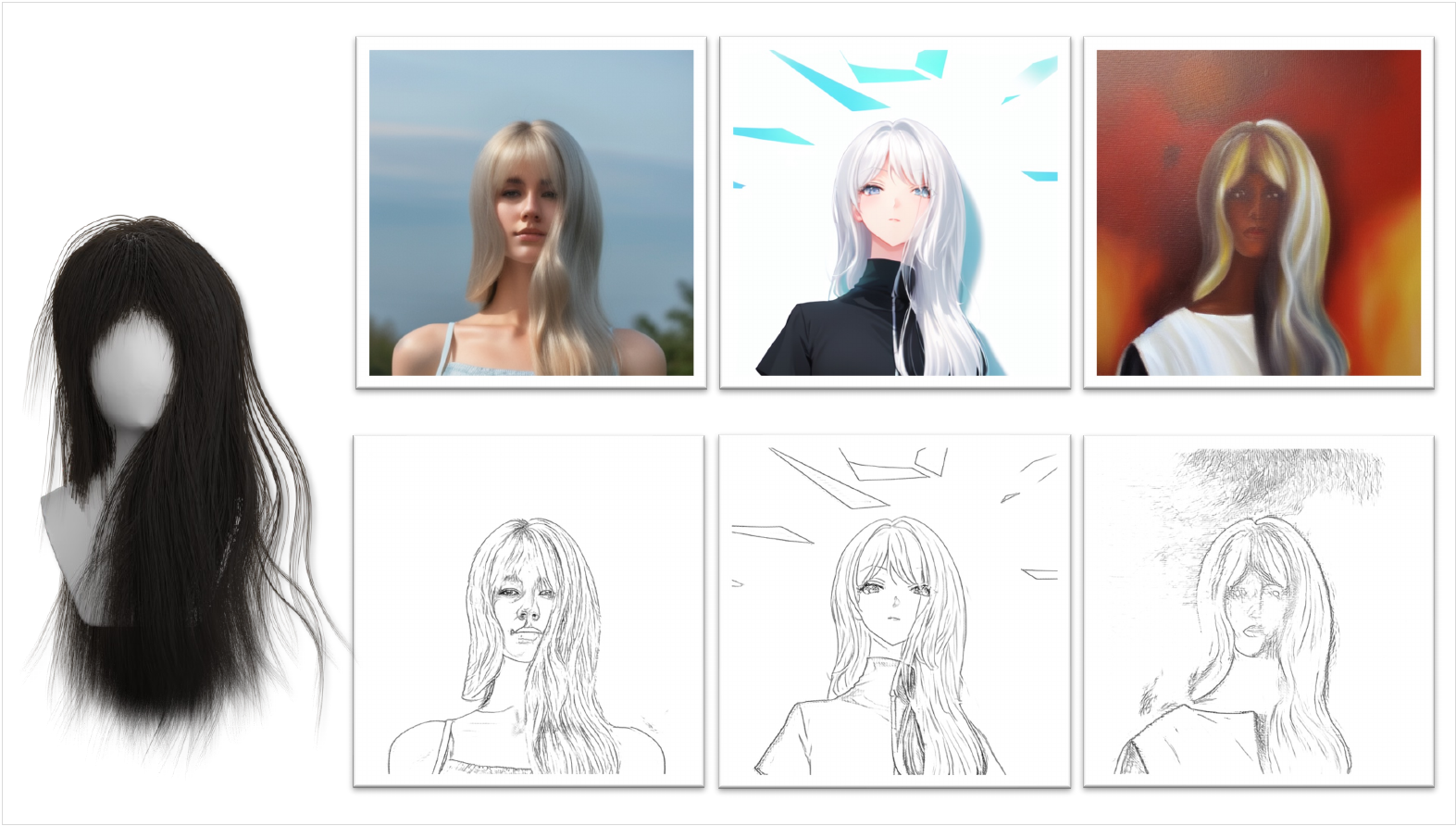}
  \caption{\textbf{Lineart extracted for various images.} For the same hairstyle under different image domains (realistic, anime and oil painting), the extracted lineart effectively captures consistent hair structure and features. }

  \label{fig:lineart}
\end{figure}

To condition our hair generation model on multi-view lineart, we first extract hair region masks from the input images using Grounding DINO~\cite{liu2025grounding} and SAM~\cite{kirillov2023segment}. This ensures that the model focuses on hairstyle-related regions, effectively removing background noise and distractions. Lineart serves as an effective representation for capturing hair strand details across various styles while mitigating interference from factors such as lighting variations, occlusions, and other irrelevant artifacts. \textcolor{black}{Fig.~\ref{fig:lineart} demonstrates that lineart extracted from the same hairstyle rendered in realistic, oil painting, and anime styles, exhibits the similar characteristic. Despite stylistic disparities, the extracted lineart retains structural coherence and granular detail fidelity, affirming its robustness to domain shifts}. We employ a lineart detector~\cite{zhang2023adding} to extract lineart from the input images, emphasizing the geometric structure and details of the hair strands.
For feature extraction, we utilize the pretrained DINOv2~\cite{oquab2024dinov2} model to extract features from the input lineart image. The resulting lineart features $\mathbf{F}$ are then concatenated and integrated into the diffusion model through a cross-attention mechanism.

To enable our model to handle image inputs from arbitrary viewpoints, we adopt a multi-view training strategy. Specifically, during training, we randomly select 1 to 8 images from the 72 multi-view image annotations for each hairstyle model, and concatenate their DINOv2 features to form the conditioning input for the diffusion model. This approach equips the model with the ability to adapt to arbitrary viewpoints and styles, enhancing the robustness and consistency of the generated hairstyles.
For inference, our method supports multi-view image input. 
As shown in Fig.~\ref{fig:comparison_qualitative}, single-view input may lack sufficient information to accurately reconstruct complex hairstyles with occluded regions. In contrast, multi-view inputs provide comprehensive details from various perspectives, enabling the model to capture both global structures and fine details, resulting in more precise and adaptable hairstyle generation across diverse styles and viewpoints.

\subsection{Parametric Braid Inpainting}

Generating realistic braided hairstyles is challenging due to their topological complexity and strand dynamics. Methods like physically-based simulators~\cite{PBSimulator, HairAnimGPU} or neural implicit surfaces~\cite{wang2021neus} often fail to maintain braid coherence.
Directly synthesizing braids via diffusion models also falters due to their reliance on weak geometric priors. 
To address this, we design an inpainting~\cite{lugmayr2022repaint} approach to seamlessly inject the braid into an already generated hairstyle. Specifically, we utilize a hybrid diffusion framework that explicitly encodes braid geometry through parametric modeling. Braids are parameterized using strand grouping templates and root guide curves, providing a structured prior. This prior is transformed into latent space and incorporated into the diffusion process using an attention-based inpainting technique, ensuring that the braid integrates naturally and cohesively with the existing hairstyle.

\vspace{4pt}
\noindent \textit{Braid Detection and UV Mapping.} 

To enable consistent braid synthesis from a reference image, our framework localizes the braid structure using a segmentation pipeline combining Grounding DINO~\cite{liu2025grounding} and SAM~\cite{kirillov2023segment}, extracting the 2D braid mask. Anatomical coherence is ensured by aligning braid roots with the scalp topology. We estimate the head pose and align a template mesh using 3DDFAv2~\cite{guo2020towards}, projecting the 2D braid mask onto the mesh’s scalp region. For each hair strand, IoU with the SAM-extracted mask identifies strands within the braid region. The strand roots determine $M_{\text{braid}}$, the braid root region in UV texture space. Manual selection of strands is also supported for defining $M_{\text{braid}}$ when the head pose estimation fails. This UV mask guides braid-specific localization and injection during the diffusion process.

\vspace{4pt}
\noindent \textit{Braid Parametrization.}
Within the braid UV mask, we select the corresponding hair strands and calculate an average curve to serve as our guide strand. Using the guide strand, we compute a Frenet--Serret frame and introduce helical patterns along the guide strand to generate the braid geometry, following~\cite{hu2014capturing}.
The shape of the braid can be adjusted by modifying parameters that control its width, thickness, and cross-sectional oscillation. 
To enhance visual smoothness and geometric fidelity, we apply Laplacian smoothing~\cite{Crane:2011:STD} to the strands, reducing high-frequency noise while preserving critical features (e.g., crossover points, torsion). The refined strands are integrated to form cohesive multi-strand braids. Finally, the parameterized hair structure is mapped into the latent space via our strand encoder $\mathcal{E}$, enabling braid topology-aware synthesis within the diffusion framework.

\vspace{4pt}
\noindent \textit{Braid Inpainting.}

Once the latent map and corresponding mask for the braid are obtained, we use an inpainting~\cite{lugmayr2022repaint} approach to inject the braid into the appropriate position on the scalp. Specifically, during denoising step of the diffusion process, the latent representations within the masked region are replaced with those from the braid’s latent map. This ensures that the braid’s geometric structure is preserved while allowing it to blend seamlessly with the surrounding hair strands.

\begin{figure}[tbp]
  \includegraphics[width=0.9\columnwidth]{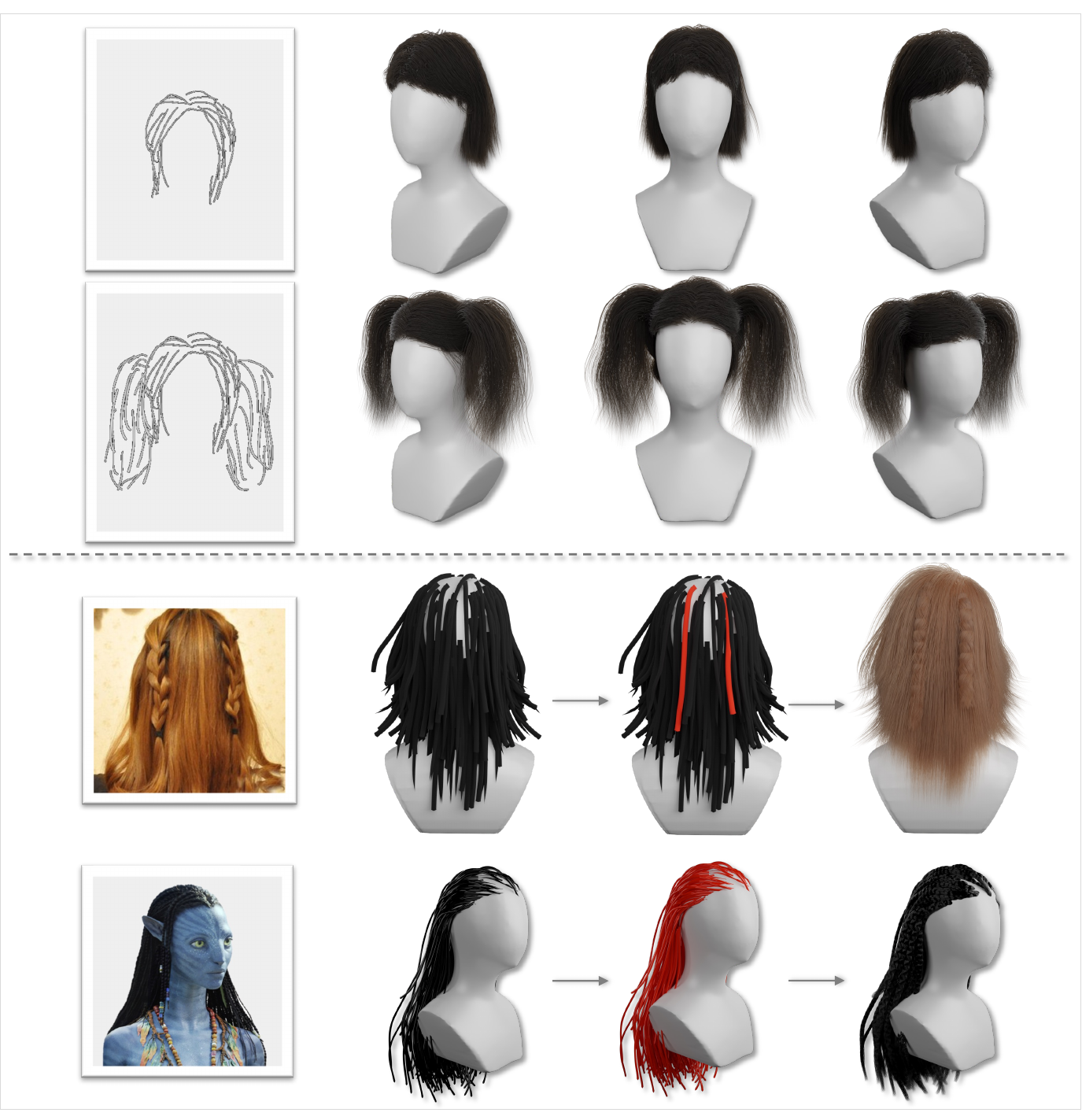}
  \caption{\textbf{Application showcase.} \textit{Row 1} show the generated hairstyles from hand-drawn sketches. \textit{Row 2} illustrate hairstyle modifications(adding pigtails) by altering specific parts in the sketches from \textit{Row 1}. \textit{Row 3-4} depict the process of generating outputs with braid using guidelines (highlighted in red).}
  \label{fig:application}
\end{figure}

%% file: TANGLED/sections/5_results.tex
\section{Results}
\label{sec:experiments}

\begin{figure*}
  \includegraphics[width=\textwidth]{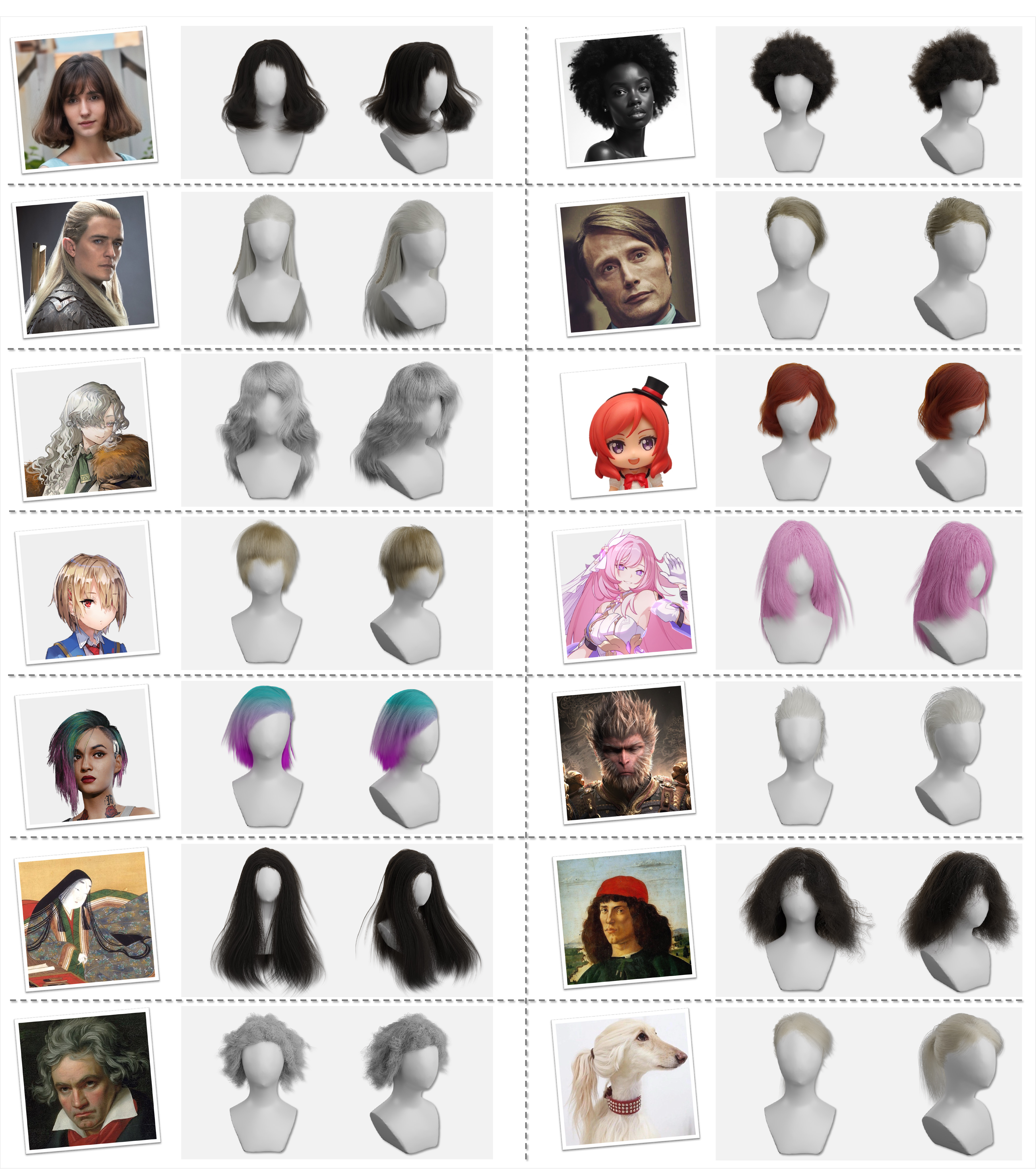}
  \label{fig:results_gallery}
  \caption{\textbf{Result gallery.} \myname can generate realistic hairstyles from image conditions with various styles, including photographs, anime, and oil paintings. For more results, please refer to the supplementary video. Note that we manually specified the color for the generated hair during the rendering process.}
  \label{fig:gallery}
\end{figure*}

We demonstrate the capability of our model to generate 3D hair with various styles from image inputs in various styles, including photographs, cartoons, and paintings, as shown in Fig.~\ref{fig:application} and Fig.~\ref{fig:gallery}. We also showcases some applications of our method. Our approach can generate hairstyles from hand-drawn sketches and allows users to modify existing lineart to create customized hairstyles, allowing users to refine hairstyles iteratively without extensive 3D expertise.

\subsection{Implementation Details}

Our diffusion model operates on input sizes of $32 \times 32$, using a U-Net~\cite{ronneberger2015u} architecture. The U-Net is configured with 8 attention heads and the channel multipliers are defined as [1, 2, 4, 4]. For training, we use the AdamW~\cite{loshchilov2017decoupled} optimizer with a learning rate of $1 \times 10^{-4}$, and follow the soft Min-SNR~\cite{hang2023efficient} weighting strategy. Noise sampling follows a cosine-interpolated density distribution. For inference, we employ DDIM~\cite{song2020denoising} sampling with 50 steps. We train our method on a single NVIDIA RTX 3090 GPU for approximately 10 days. Classifier-Free Guidance~\cite{ho2022classifier} is applied during training with a probability of 0.1 to randomly mask the image condition.

\subsection{Comparisons}

We evaluate our method against state-of-the-art 3D hair generation/reconstruction approaches: HAAR~\cite{HAAR:CVPR:2024}, a text-guided generative model, and HairStep~\cite{zheng2023hairstep}, an optimization-based reconstruction method that currently achieves state-of-the-art performance. For benchmarking, we use a 5\% test split of our MultiHair Dataset, ensuring no overlap with training data.
HAAR generates hairstyles from text descriptions, requiring a Visual Question Answering (VQA) system to convert input images into textual representations via BLIP~\cite{li2022blip}. HairStep, in contrast, directly optimizes 3D strands from single-view inputs. To ensure a fair comparison, we use the official pre-trained models for both baselines, avoiding implementation biases.

\vspace{4pt}
\noindent \textit{Quantitative Comparisons.}

We conduct a quantitative comparison to evaluate generation quality (geometric and semantic fidelity) and computational efficiency against state-of-the-art methods. Three metrics are employed: Point Cloud IoU to measure geometric similarity between generated 3D hair and ground-truth point clouds; the CLIP Score~\cite{radford2021learning} to evaluate semantic alignment between rendered images and input conditions; and the Chamfer Distance for quantifying geometric accuracy by comparing nearest-neighbor distances between generated and ground-truth surfaces.

For HAAR and our method, we generate 10 stochastic samples per test case to account for generative variability and report mean metric values to ensure statistical robustness. As shown in Tab.~\ref{tab:comparison-quantitative}, our method surpasses other approaches, demonstrating its capability to generate superior results with multi-view and multi-style image inputs.

\begin{figure}[tbp]
  \includegraphics[width=\columnwidth]{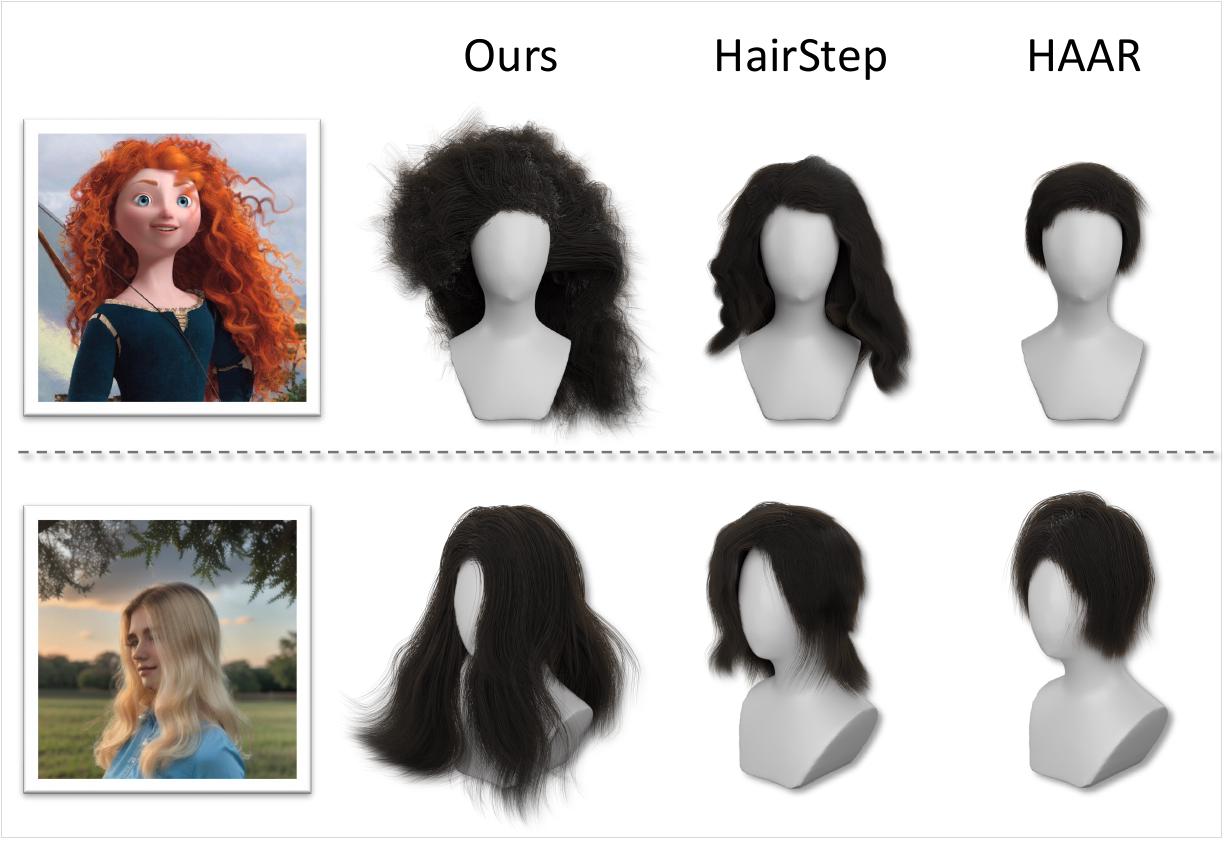}
  \caption{\textbf{Qualitative comparison.} For various input images, our method produces more aligned and detailed hairstyles compared to HairStep and HAAR. For more comparison, please refer to Fig~\ref{fig:comparison_qualitative2}.}
  \label{fig:comparison_qualitative}
\end{figure}

\vspace{-4pt}
\begin{table}
  \small
  \caption{\textbf{Quantitative comparisons and evaluations.} Ours($\mathbf{R}$) refers to our method trained on real images. Ours($\mathbf{V_1}$) and Ours($\mathbf{V_4}$) refer to our method using randomly selected one-view and four-view inputs. We exclude test cases where HairStep failed to optimize hair strands, which is in favor of the HairStep. Yet, our approach achieves the best performance regardless.}
  \label{tab:comparison-quantitative}
    \begin{tabular}{cccccc}
    \toprule
    Methods        & Clip Score $\uparrow$ & CD($\times10^{-2}$m)$\downarrow$ & IoU $\uparrow$ \\
    \toprule
    HairStep     &  69.09   &  0.0112   &  52.78\%  \\
    HAAR     &  73.81   &  0.0336   &  42.14\%   \\
    \midrule
    Ours($\mathbf{R}$)        &  75.81   &  0.0051   &  45.00\% \\
    \midrule
    Ours($\mathbf{V_1}$)   &  76.77   &  0.0039   &  53.06\% \\
    Ours($\mathbf{V_4}$)        &  79.04   &  0.0033   &  53.69\% \\
    \bottomrule
    \end{tabular}
\end{table}

\vspace{4pt}
\noindent \textit{Qualitative Comparisons.}
\label{sec:qualitative_comp}
To further assess performance, we conduct qualitative comparisons across diverse input styles and viewpoints. As illustrated in Fig.~\ref{fig:comparison_qualitative}, our method outperforms baselines in preserving fine geometric details (e.g., curls, braids, split ends) and maintaining semantic alignment with input conditions (e.g., bangs, parting, volume).
HAAR’s reliance on a VQA system to extract textual descriptions from images introduces semantic ambiguities, often yielding hairstyles misaligned with input intent. HairStep, optimized for frontal-view reconstruction, struggles with viewpoint invariance — occluded regions or non-frontal inputs.

\begin{figure}[thbp]
  \includegraphics[width=\columnwidth]{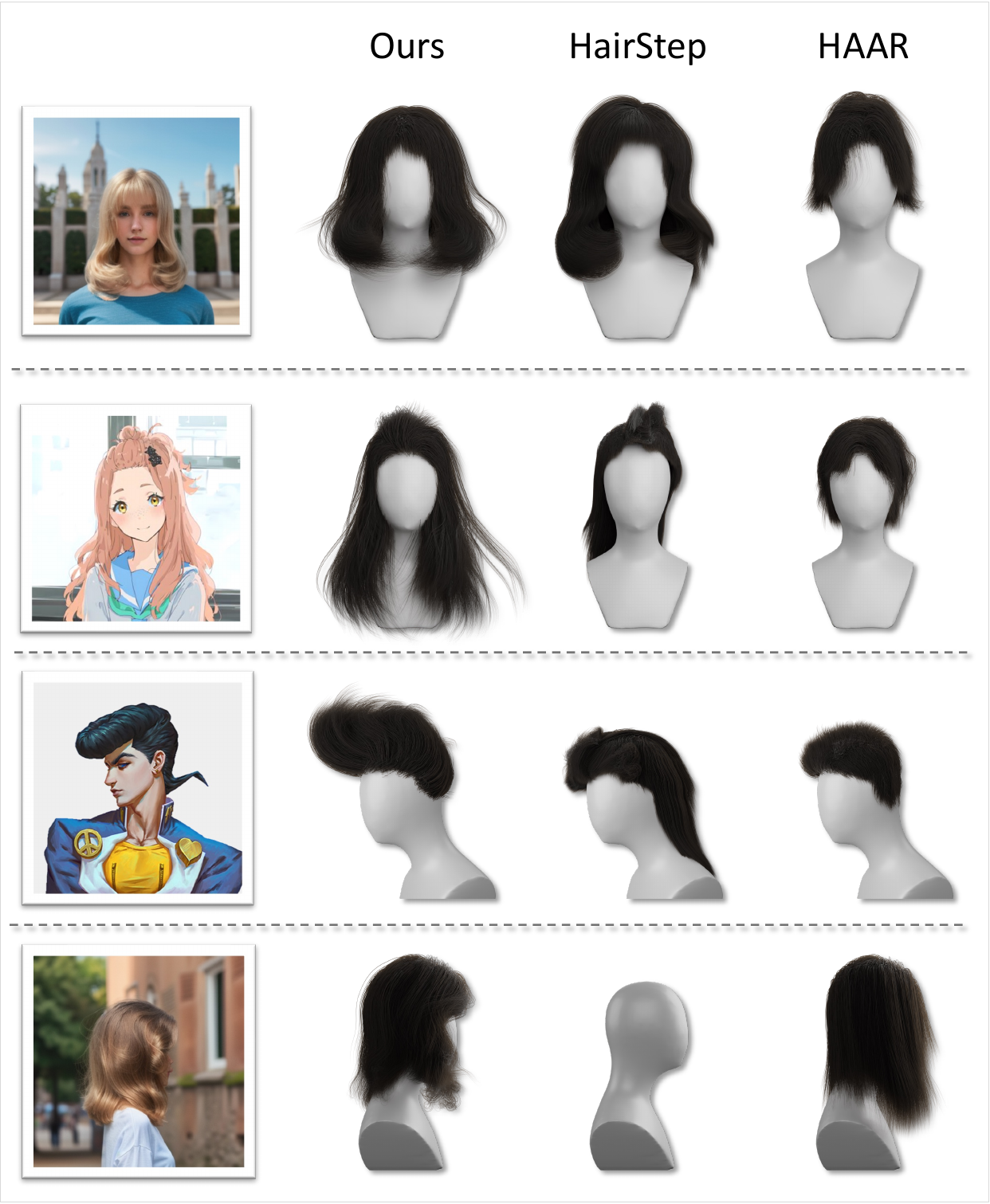}
  \caption{\textbf{Qualitative comparison.} Our model accurately captures hairstyle structure and details, demonstrating superior adaptability to diverse input conditions. Notice that HairStep fails in cases where the face is not recognizable, as it cannot optimize the hairstyle without facial guidance(\textit{Row 4}). While HAAR relying on textual descriptions extracted from the image, struggles to preserve fine details.}
  \label{fig:comparison_qualitative2}
\end{figure}

\begin{figure}[thbp]
  \includegraphics[width=\columnwidth]{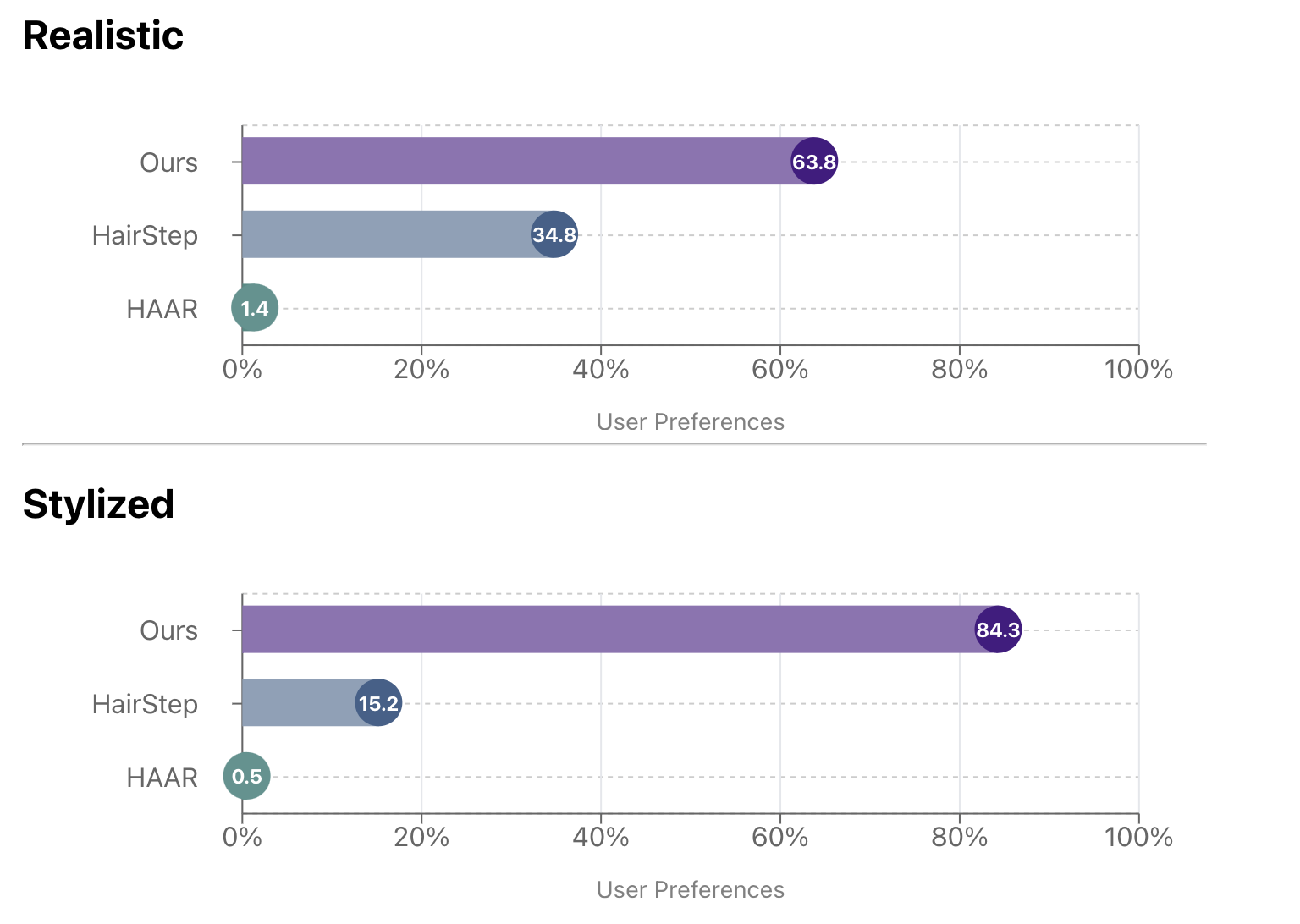}
  \caption{\textbf{User study result.} The preferences of 42 participants are illustrated for two input categories: Realistic (\textit{top}) and Stylized (\textit{bottom}). For realistic inputs, our method received the highest user preference (63.8\%), significantly outperforming HairStep (34.8\%) and HAAR (1.4\%). Similarly, for stylized inputs, our method is strongly favored (84.3\%) compared to HairStep (15.2\%) and HAAR (0.5\%).  }

  \label{fig:user-study}
\end{figure}

\vspace{4pt}
\noindent \textit{User Study.}

Additionally, we conduct a user study with 42 participants across 10 test cases spanning diverse input styles and viewpoints. For each case, we generate viewpoint-aligned renderings from our method and two baselines (HAAR and HairStep), ensuring direct visual comparability. Participants evaluated outputs based on structural fidelity (hairstyle geometry), detail retention (strand-level features), and stylistic coherence (alignment with input conditions). As shown in Fig.~\ref{fig:user-study}, our method achieved 63.8\% preference in realistic scenarios and 84.3\% in stylized cases, outperforming baselines by significant margins.

\begin{figure}[tbp]
  \includegraphics[width=\columnwidth]{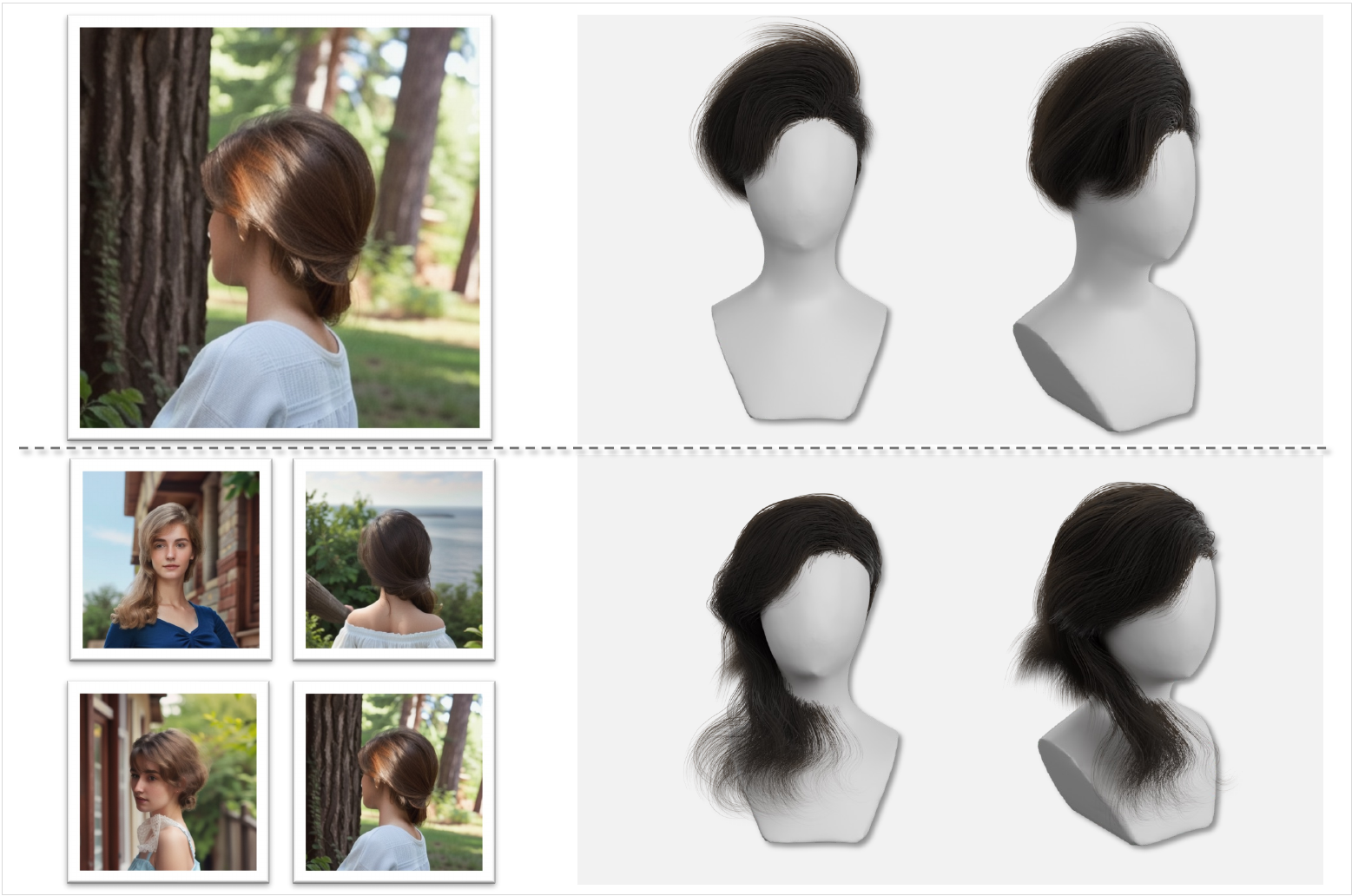}
  \caption{\textbf{Generated results using single-view and multi-view inputs.} With single-view input(\textit{Row 1}), the asymmetrical shoulder-length hair on the occluded side cannot be accurately reconstructed, resulting in missing details. In contrast, multi-view inputs (\textit{Row 2}) enable the model to capture the full structure of the hairstyle, including the previously occluded regions }
  \label{fig:multiview}
\end{figure}

\subsection{Evaluation}
To evaluate the effectiveness of our lineart extraction and multi-view conditioning approach, we conducted a series of experiments. 
As shown in Fig.~\ref{fig:multiview}, single-view image conditions may fail to generate complex hairstyles where parts of the hairstyle are occluded. When multiple views are provided, the model can accurately reconstruct the hairstyle with improved precision and detail.

Furthermore, as presented in Table~\ref{tab:comparison-quantitative}, we compared the performance of our method under two conditions: using a randomly selected single-view image and using four randomly selected views as input conditions on the test set. The results demonstrate that our method generates significantly more accurate hairstyles when conditioned on four views, highlighting the advantages of multi-view conditioning for capturing both global structures and fine-grained details of complex hairstyles. We also compared the performance of our method training on lineart condition and generated image condition. The results show that the model trained on lineart outperforms the one trained on rendered images across all metrics. This indicates that training with lineart enables the model to better capture the structure and features of hairstyles, leading to superior performance when generating hairstyles from diverse styles and viewpoints in the image conditions.

%% file: TANGLED/sections/6_conclusion.tex
\section{Discussions}
\label{sec:conclusions}

We have introduced TANGLED, a novel multi-view lineart-conditioned diffusion model for 3D hair strand generation. We present the MultiHair dataset, a diverse collection that expands hairstyle representation with underrepresented textures and complex geometries. Our diffusion framework conditioned on multi-view linearts enables flexible and accurate generation across various styles and viewpoints. Furthermore, our parametric post-processing refines braid-specific constraints, enhancing the coherence of intricate styles.
While TANGLED achieves significant advances, challenges remain. First, the MultiHair dataset, though more diverse, still lacks the capacity to model ultra-high-frequency strand details. Second, our braid generation pipeline struggles with extreme head pose, i.e., yaw/pitch \textgreater 75°, due to anchor point occlusion. Lastly, pixel-level alignment between generated hair and input images is limited, which is largely constrained by dataset size. Future work will focus on expanding dataset coverage, improving pose estimation, and enhancing alignment precision. These efforts will further TANGLED’s potential to create realistic, culturally inclusive 3D hairstyles for diverse digital applications.